\documentclass{article}
    \PassOptionsToPackage{numbers, compress}{natbib}

\usepackage[final]{neurips_2020}

\usepackage[utf8]{inputenc} 
\usepackage[T1]{fontenc}    
\usepackage{hyperref}       
\usepackage{url}            
\usepackage{booktabs}       
\usepackage{amsfonts}       
\usepackage{nicefrac}       
\usepackage{microtype}      
\usepackage{graphicx}
\usepackage[ruled,vlined]{algorithm2e}

\usepackage{color}
\title{Compositional Prototype Network with Multi-view Comparision for Few-Shot Point Cloud Semantic Segmentation}

\author{
 Xiaoyu Chen\textsuperscript{1}, Chi Zhang\textsuperscript{2}, Guosheng Lin\textsuperscript{2}, Jing Han\textsuperscript{1}, \\\textsuperscript{1} Nanjing University of Science and Technology \\ \textsuperscript{2} Nanyang Technological University \\
 115104000466@njust.edu.cn
} 

\begin{document}

\maketitle

\begin{abstract}
Point cloud segmentation is a fundamental visual understanding task in 3D vision. A fully supervised point cloud segmentation network often requires a large amount of data with point-wise annotations, which is expensive to obtain. In this work, we present the Compositional Prototype Network that can undertake point cloud segmentation with only a few labeled training data. Inspired by the few-shot learning literature in images, our network directly transfers label information from the limited training data to unlabeled test data for prediction. The network decomposes the representations of complex point cloud data into a set of local regional representations and utilizes them to calculate the compositional prototypes of a visual concept. 
Our network includes a key Multi-View Comparison Component that exploits the redundant views of the support set. To evaluate the proposed method, we create a new segmentation benchmark dataset,  ScanNet-$6^i$, which is built upon ScanNet dataset. Extensive experiments show that our method outperforms baselines with a significant advantage. Moreover, when we use our network to handle the long-tail problem in a fully supervised point cloud segmentation dataset, it can also effectively boost the performance of the few-shot classes.

\end{abstract}

\section{Introduction}

With the advances of 3D sensors, visual understanding of 3D data with deep learning has attracted broad interests in recent years. Point cloud segmentation is one of the most fundamental 3D vision tasks that aims to predict the label of each point in the space.
However, training a deep model for point cloud segmentation requires a large amount of point-wise annotated data, which is usually very expensive to obtain.

\begin{figure}
  \centering
  \includegraphics[width=7.8cm]{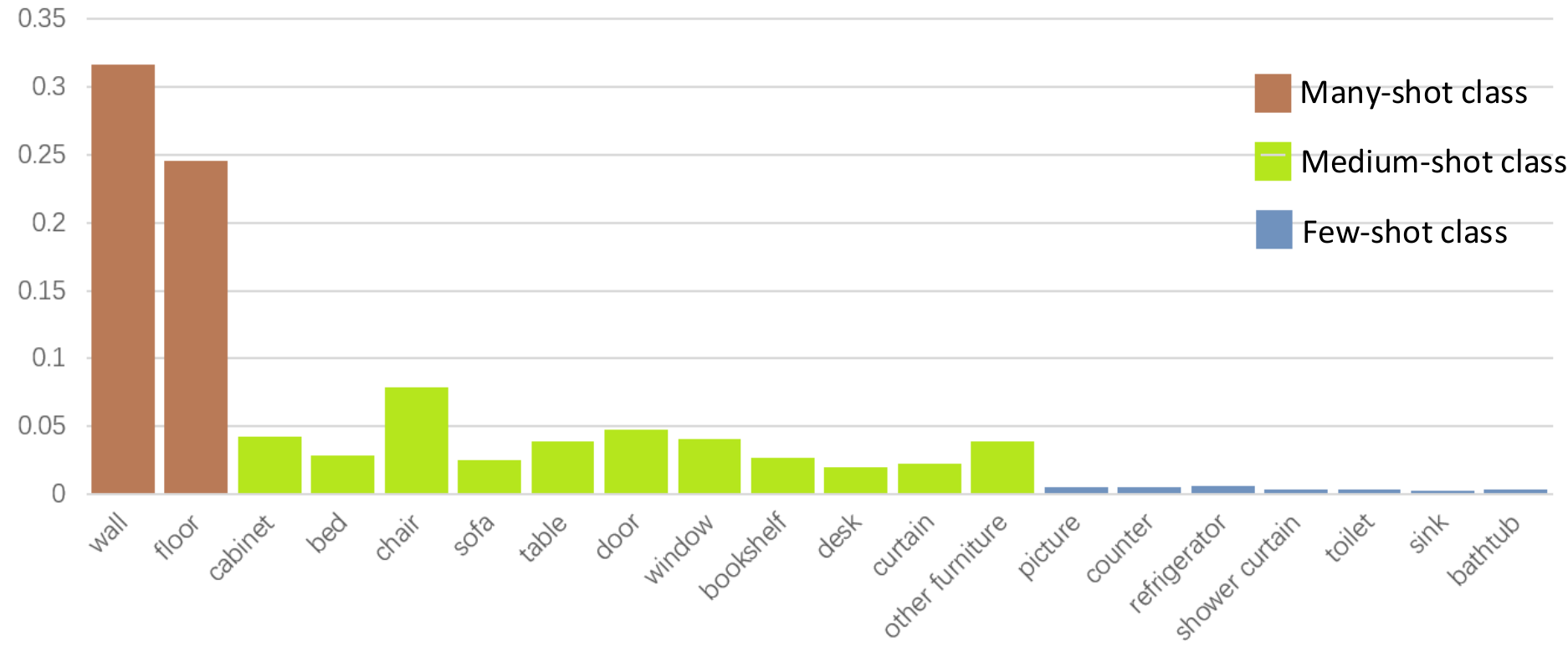}
  \caption{The dataset statistics in ScanNet dataset\citep{dai2017scannet}. We plot the proportions of labeled points in different classes. Point cloud datasets usually show significant data imbalance.
  }\label{fig:scannet_prop}
\end{figure}

Few-shot learning is a widely studied technique to handle the data-hungry problem of deep models, which targets at an extreme case where only a few labeled data is available for the prediction of  a new class. In this paper, we study few-shot learning for point cloud segmentation, where only a few labeled  data is available to undertake 3D point cloud segmentation.
Few-shot learning in the point cloud segmentation task has many practical values. First, since the point cloud data is expensive to annotate, the few-shot learning model can largely alleviate the efforts for it. Second, as current point cloud datasets are highly imbalanced, the few-shot classes with scarce training data may not perform well under the fully supervised training paradigm.
 For example, the popular 3D point cloud dataset  ScanNet \citep{dai2017scannet}  only contains 1,513 labeled data examples, which is  much fewer than the number in image datasets. Moreover, the severe data imbalance problems also widely exist in point cloud datasets. As the statistics show in Fig.\ref{fig:scannet_prop}, the labeled points in two many-shot classes account for more than fifty percent of the whole dataset, while the points in few-shot classes can be  lower than three percent.
All these issues above motivate us to study the few-shot point cloud segmentation, which emphasizes on data efficiency for point cloud segmentation.

In our work, we first examine a collection of few-shot learning techniques in images as baselines for point cloud data, such as metric-based methods \citep{Sung_2018_CVPR,wang2019panet} and relation-based methods~\cite{liu2020crnet,zhang2019canet,zhang2019pyramid}, which are two dominant  streams in few-shot image segmentation.
Metric-based methods usually use a global vector embedding to represent a visual concept, which is called a prototype, and employ a  distance metric to determine the pixel-level similarity for classification. On the other hand, relation-based methods use a deep network to implicitly transfer the prototype information from training data into the unlabeled data for prediction. 

We find that if we directly apply these techniques to point cloud data, the results are often not ideal.
The reason we find is that compared with image data, the point cloud data contain more structural information and regional information. If we directly pool all the points in the space to a single vector representation, much useful discriminative information and spatial information are lost, resulting in inferior performance.
Based on such intuition, we propose to decompose the data with complex structural representations into local regional representations and propose a Multi-View Comparison Component (MVC) to transfer label information directly at region-levels. In our proposed MVC component, the regional representations from training data are directly transformed to the weights of the Convolutional kernels to make help the predictions on  unlabeled testing data. It can effectively utilize local discriminate information to reason the label of each point in the space with redundant views of embeddings.

To evaluate the effectiveness of our proposed few-shot learning algorithm in point cloud segmentation, we create a new benchmark ScanNet$-6^i$ for few-shot point cloud segmentation.
We also investigate the proposed few-shot learning model in handling the class-imbalance problems in a fully supervised setting. Our method can also effectively enhance the performance of the fully supervised segmentation model, particularly on the few-shot classes.
The contributions in this paper  are summarized as follows:
1) We propose a novel framework for few-shot point cloud segmentation based on  multi-view comparison ;
2) We investigate many previous methods for few-shot image segmentation to handle few-shot point cloud segmentation as baselines;
3)To evaluate the performance of our proposed network, we create a new benchmark  dataset ScanNet$-6^i$ built upon the ScanNet dataset\citep{dai2017scannet}.
4) We demonstrate that our few-shot segmentation model can also be used to handle the data imbalance problem in the fully supervised point cloud segmentation task.

\section{Related Works}

\subsection{Few-shot Semantic Segmentation in 2D images}
Previous works \citep{Sung_2018_CVPR,wang2019panet,liu2020crnet,zhang2019canet,zhang2019pyramid} in few-shot image segmentation often employs a two-branch design. The support and query images are fed to encoder networks, and the foreground features of the support set are collected to compute the class-wise prototypes. Then the query points are classified in the metric space based on the prototypes with various metric models.
For example, in \citep{shaban2017one}, the support images and masks are concatenated and then fed into a conditioning branch (VGG) to obtain the prototype kernel. The query images are fed into the segmentation branch (FCN-32s) to generate point embeddings, matching the prototype kernel based on Convolution.
In PANet \citep{wang2019panet} and CANet \citep{zhang2019canet}, support and query branches share the encoder network, and the foreground prototype is computed using global average pooling on the foreground features. Then the foreground prototype is used to match the foreground part in the query images with cosine distances and Relation Network \citep{Sung_2018_CVPR}.
 PGNet \citep{Zhang_2019_ICCV} models structured segmentation data with graphs and apply attentive graph reasoning to propagate label information from support data to query data. The attentive graphs help to compute more accurate prototypes.

\subsection{Deep Metric Learning}
Metric learning methods attempt to map data to an embedding space, where data with similar characteristics are close and dissimilar data are apart.
Metric learning is widely used in tasks with fewer annotated data, such as video object tracking and few-shot recognition. The metric learning is also used as an alternative to softmax in classifiers\citep{Sung_2018_CVPR}.

DeepNCM \citep{guerriero2018deep} directly learn highly non-linear deep representations of data based on metric learning, and the DeepNCM are proved to alleviate class-imbalance effect in \citep{kang2019decoupling}. 
The conventional metric models in metric learning are often simple and have an explicit mathematical definition, e.g., Euclidean distance, Cosine distance and Earth Mover's distance \citep{snell2017prototypical, vinyals2016matching,zhang2020deepemd}. Relation Network \citep{Sung_2018_CVPR} propose a data-driven metric algorithm to learn metrics end-to-end from the data. In online object tracking, Inner-Product is widely used to build the relation between annotated targets and searching images \citep{bertinetto2016fully}. SiamMask further \citep{wang2019fast}  integrates one-shot segmentation into the tracker to enhance the performance.
More recently, DeepEMD~\citep{zhang2020deepemd} employs the earth mover's distance to compute a structural distance between image regions for few-shot image classification.

\subsection{3D Point Cloud Analysing}
With the release of large scale point cloud datasets, many methods are proposed to analyze such unordered and sparsely distributed data. There are mainly three categories in analysis methods for 3D point cloud: point-based methods \citep{qi2017pointnet,qi2017pointnet++,thomas2019kpconv}, multi-view methods \citep{jaritz2019multi}, and volumetric methods \citep{3DSemanticSegmentationWithSubmanifoldSparseConvNet}.

Point-based models adopt new-designed Convolution operators to tackle the disordered and sparse data of point clouds. They extract features from each point and then obtain the global representation. Multi-view methods recognize 3D objects from a collection of their rendered views on 2D images and convert 3D tasks to 2D tasks. One of the significant advantages of Multi-view methods is that the priors or knowledge of 2D images can be exploited.

Volumetric based methods are also widely explored, where the unordered and sparse point clouds are put into volumetric grids. As the voxelized point clouds are sparsely distributed, simply using standard 3D Convolution on such data are low efficiency and have high memory cost. Submanifold Sparse Convolutional Networks (SparseConvNet) \citep{3DSemanticSegmentationWithSubmanifoldSparseConvNet} design a new series of sparse Convolutional operators to process spatially-sparse data. The SparseConvNet is more efficient and makes it possible to process fine-grained voxelized point clouds.

\section{Problem Set-up}
Suppose that our model is trained on annotated point clouds with the class set $\mathcal{C}_{train}$, and the goal is to make predictions on point clouds with the new class set $\mathcal{C}_{test}$, where only a few annotated data is available. At both of training and testing periods, the examples providing annotated masks in the forward pass are called support set $\mathcal{S} = \{(x_s^i,y_s^i(c))\}^{k}_{i=1}$, where $x_s^i \in \mathbb{R}^{H_i\times W_i\times 6}$ is a point cloud (containing XYZ and RGB data) and $y_s^i(c) \in \mathbb{R}^{H_i\times W_i}$ is binary masks of class $c$. The $k$ is the example number of the support set in $k$-shot learning. And those providing labels for computing loss or evaluation are called query set $\mathcal{Q}=\{x_q,y_q(c)\}$ where $x_q$ is the query cloud and $y_q(c)$ is the ground-truth mask for class $c$.
Following the training paradigm in few-shot image segmentation literature~\citep{zhang2019canet}, our model is trained by episodically sampling paired examples from the class set $\mathcal{C}_{train}$ and evaluated by sampling paired examples from the class set $\mathcal{C}_{test}$

\section{Method}
\begin{figure}
  \centering
  \includegraphics[height=4.3cm]{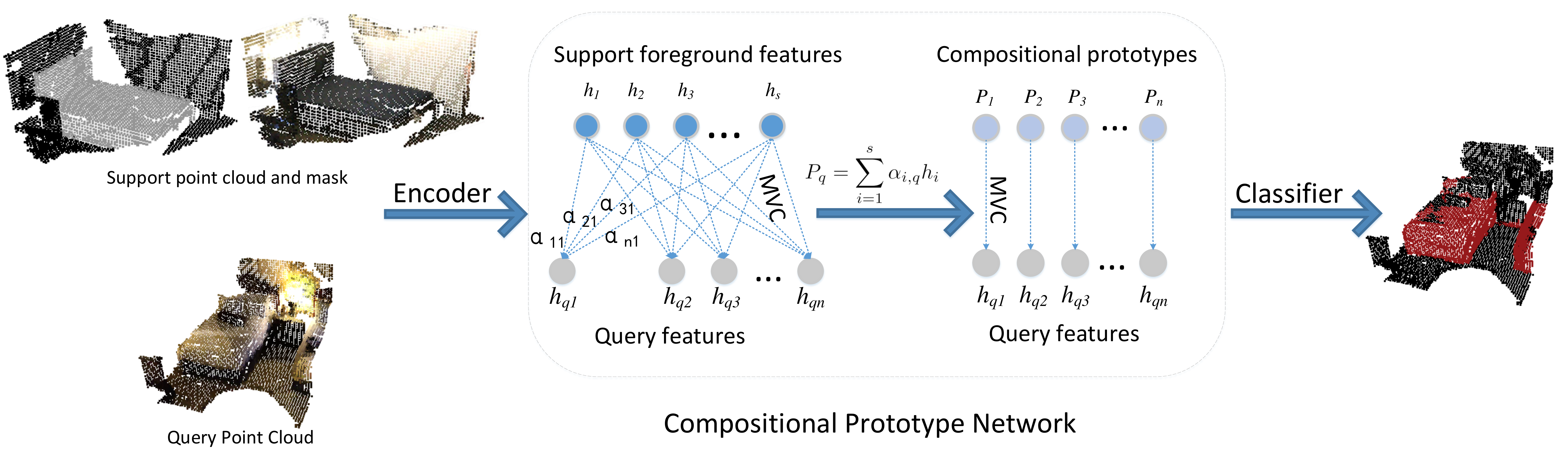}
  \caption{The network structure for few-shot point cloud segmentation. Blue dotted arrows indicate the Multi-view Comparison Component (MVC), the foreground features are structured as compositional prototypes to evaluate if the query points are foreground points.}
  \label{fig:stru}
\end{figure}

In this section, we present our framework for few-shot point cloud segmentation, which includes  the key  Multi-view Comparison Component.  The proposed network consists of an encoder, a Compositional Prototype Network, and a classifier. The encoder is to represent support and query data in a shared embedding space, and the Compositional Prototype Network is to generate region aligned prototypes. Our network structure is shown in Fig.~\ref{fig:stru} and the whole training and testing procedures on few-shot segmentation are summarized in Algorithm~\ref{algo}.

\begin{algorithm}[t]
\SetAlgoLined
\SetAlgoHangIndent{12pt}
\SetKwInOut{Input}{Input}\SetKwInOut{Output}{Output}
\Input{A training set $\mathcal{D}_{train}$ and a testing set $\mathcal{D}_{test}$}
\For{each episode $(\mathcal{S}_i, \mathcal{Q}_i)\in \mathcal{D}_{train}$}
{
1, Extract support foreground features ${h_s}$ and query features from the support set $\mathcal{S}_i$ and query set $\mathcal{Q}_i$ with the Encoder Network. \\
2, Construct regional foreground features ${h_i}$ from support foreground features ${h_s}$. \\
3, Estimate the Compositional Prototype ${P_q}$ for each query features based on the Eqn .\ref{proto}. \\
4, Predict the segmentation probabilities and masks for the query points based on the Multi-view Convolution. \\
5, Compute the loss. \\
6, Compute the gradient and optimize via SGD.
}

\For{each episode $(\mathcal{S}_i, \mathcal{Q}_i)\in \mathcal{D}_{test}$}
{
1, Extract support foreground features ${h_s}$ and query features from the support set $\mathcal{S}_i$ and query set $\mathcal{Q}_i$ with the Encoder Network. \\
2, Construct regional foreground features ${h_i}$ from support foreground features ${h_s}$. \\
3, Estimate the Compositional Prototype ${P_q}$ for each query features based on the Eqn .\ref{proto}. \\
4, Predict the segmentation probabilities and masks for the query points based on the Multi-view Convolution. 
}
  \caption{Training and evaluating the Compositional Prototype Network.}\label{algo}
\end{algorithm}

\subsection{Multi-view Comparison Component for Metric Learning}

Various metric algorithms are used for metric measurements such as Euclidean distances and Cosine similarities. These similarities are computed directly between embeddings. With the advances of a deep neural network, a data-based similarity model, Relation Network \citep{Sung_2018_CVPR}, is proposed.

In Relation Network, relations are computed with weights and bias, which is independent of input features. This process is based on the transformation and fusion of the embeddings and is not a direct way to obtain relations. Here, we present our Multi-view Comparison Convolution ($MConv$) to generate multi-view kernels from support set and predict similarities by convolving query embeddings.

Taking $h_{s}$ as support embeddings and $h_{q}$ as the query embeddings. The multi-view kernels of the support set are generated with the $h_{s}$ by:

\begin{equation}\label{tconv_w}
k_s = \phi(h_{s}) = w_c (w_l h_{s})^T + b_k,
\end{equation}

where the $w_l$ is used to adjust the feature-length of $h_{q}$, $w_c$ is to increase the kernels in $MConv$, and $b_k$ is the bias term. The various kernels indicate the different views of $MConv$. These parameters are updated by end-to-end training. Then the relation between $h_i$ and $h_j$ is computed by:

\begin{equation}\label{tconv}
g_{s,q} = Conv(h_q | k_s) = h_{q} * k_s + b,
\end{equation}

where $k_i$ is used as kernel, and $b$ is the bias term.

\begin{figure}
  \centering
  \includegraphics[width=12cm]{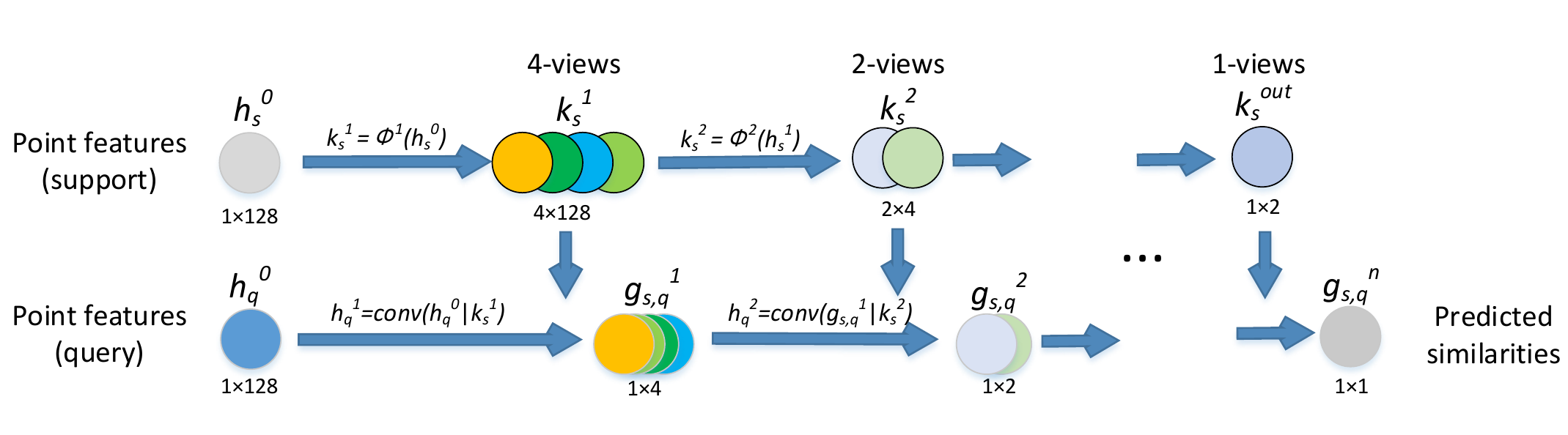}
  \caption{The graphical demonstration for Multi-view Comparison Component (MVC) with multiple layers. The support features are computed recursively as the convolution kernels for query features matching.}
  \label{fig:tconv}
\end{figure}

The multi-view kernels and output features are used as a new support and query features of the next layer respectively to construct multiple layers of $MConv$, shown in Eqn .\ref{tconvs}. The similarities between support and query points can be computed recursively between themselves. The graphical demonstration for Multi-view Comparison Component with multiple layers is shown in Fig. \ref{fig:tconv}

\begin{equation}\label{tconvs}
g_{s,q} = MConv(h^{n-1}_q | k^n_s) = h^{n-1}_q * k^n_s + b^n.
\end{equation}

The views of $MConv$ decide the number of output channels, and the view number of the output layer is set to 1 to obtain scalars of similarities.

\subsection{One-Shot Point Cloud Semantic Segmentation with Compositional Prototypes}

PGNet \citep{Zhang_2019_ICCV} was proposed to use attentive graphs to model structured segmentation data. Instead of implement global average pooling on foreground features of support set, PGNet adopts graph attention unit(GAU) to computing the Inner-Product relation between the regional support features and query features for prototype alignment. We improve the GAU by replacing the Inner-Product operators with the proposed Multi-view Comparison Component, where support and query features are recursively computed for similarities between query embeddings and regional support embeddings according to Eqns. \ref{tconv_w}  \textasciitilde \ref{tconvs}.

Taking $h_{i}$ as regional support embeddings, and $h_{q}$ as the target embeddings, Their similarities $g_{s,q}$ are used to compose the regional features for each query features by:

\begin{equation}\label{proto}
P_q=\sum\limits_{i=1}^{s} {\alpha_{i,q}  h_{i}}=\sum\limits_{i=1}^{s} {\frac{exp(g_{i,q})}{\sum\limits_{j=1}^{s}{exp(g_{j,q})}}  h_{i}}.
\end{equation}

Then, the compositional prototypes are used to build a metric space, where query points are classified by comparing with the prototypes based on Multi-view Comparison Component. We also add five Convolution layers at the end as the post-processing part.
\begin{figure}
  \centering
  \includegraphics[width=12cm]{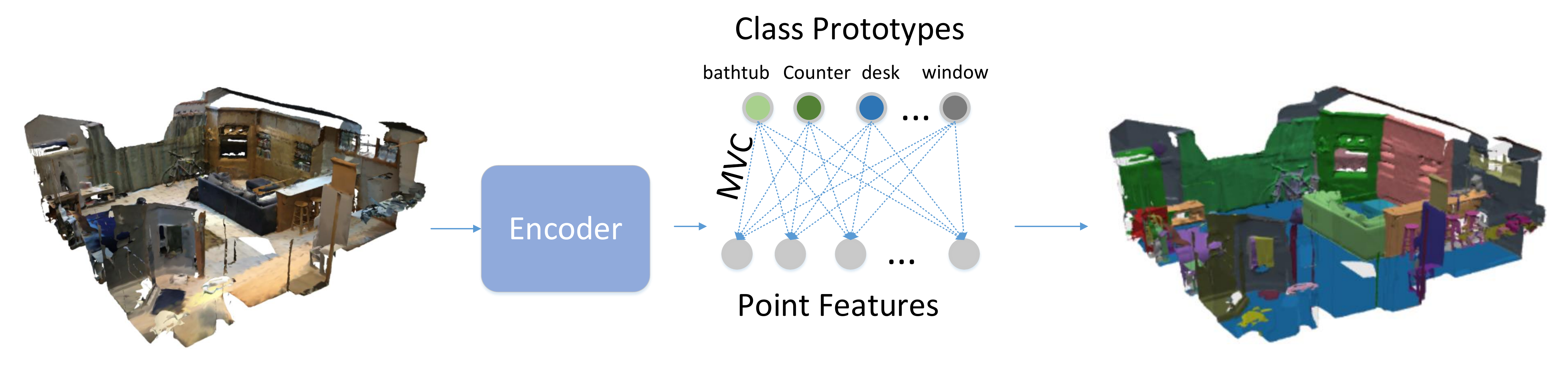}
  \caption{Structure of metric-based supervised point cloud segmentation. Blue dotted arrows indicate the Multi-view Comparison Component. The randomly initialized prototypes are used to evaluate the correlation between classes and query points for classification.}
  \label{fig:supervised}
\end{figure}
\subsection{Comparison of Inner-Product, Convolution and Multi-view Comparison Convolution in Metric Learning}

Inner-Product is a simple operation to compute correlations between equal-length vectors by accumulating the products of the corresponding entries. Inner-product exists in Convolution, which is to calculate between the Convolution kernels and concatenated inputs. The kernels are independent of inputs, and it is not a direct way to compare the inputs. Still, the multiple Convolution layers are able to extract deeper features of inputs.

Multi-view Comparison Convolution ($MConv$) improves the Inner-Product operations to a more generalized way to compare. The multi-view embeddings of the support set are generated to measure the similarities from redundant views. The learnable parameters are only used to adjust support embeddings and extend views. It is a direct method to compare and also easy to be extended to multiple layers. Besides, Compared with Convolution, $MConv$ saves much computation costs as it does not need to concatenate inputs for Convolutions.

\subsection{Metric Learning in Supervised Point Cloud Segmentation}\label{sup}

As strongly class-imbalance exists in point cloud dataset, shown in Table. \ref{fig:scannet_prop}, the supervised methods of point cloud segmentation often have poorer performance in few-shot classes. 

We design a metric-based method, shown in Figure \ref{fig:supervised}, which modify the classifier of the supervised methods. We create prototypes for each class and compare the embeddings from encoder with the prototypes for classification. The Multi-view Comparison Component is used to compute the similarities between the point embeddings and prototypes. Both of the prototypes and Multi-view Comparison components are updated by end-to-end training.


We argue that points belong to the same have similar feature embeddings, which are close to their prototypes. We can also consider the last Convolution of the classifier as a metric operation where prototypes are the kernel, and the similarities are computed based on Inner-product. In our method, the Multi-view Comparison Component can learn deeper features of similarities and make the metric space more flexible. Our metric-based method makes the model to learn to find the prototypes of each class and obtain more balanced representations in the metric space.

\section{Dataset and Metric}
\begin{figure}
  \centering
  \includegraphics[width=10cm]{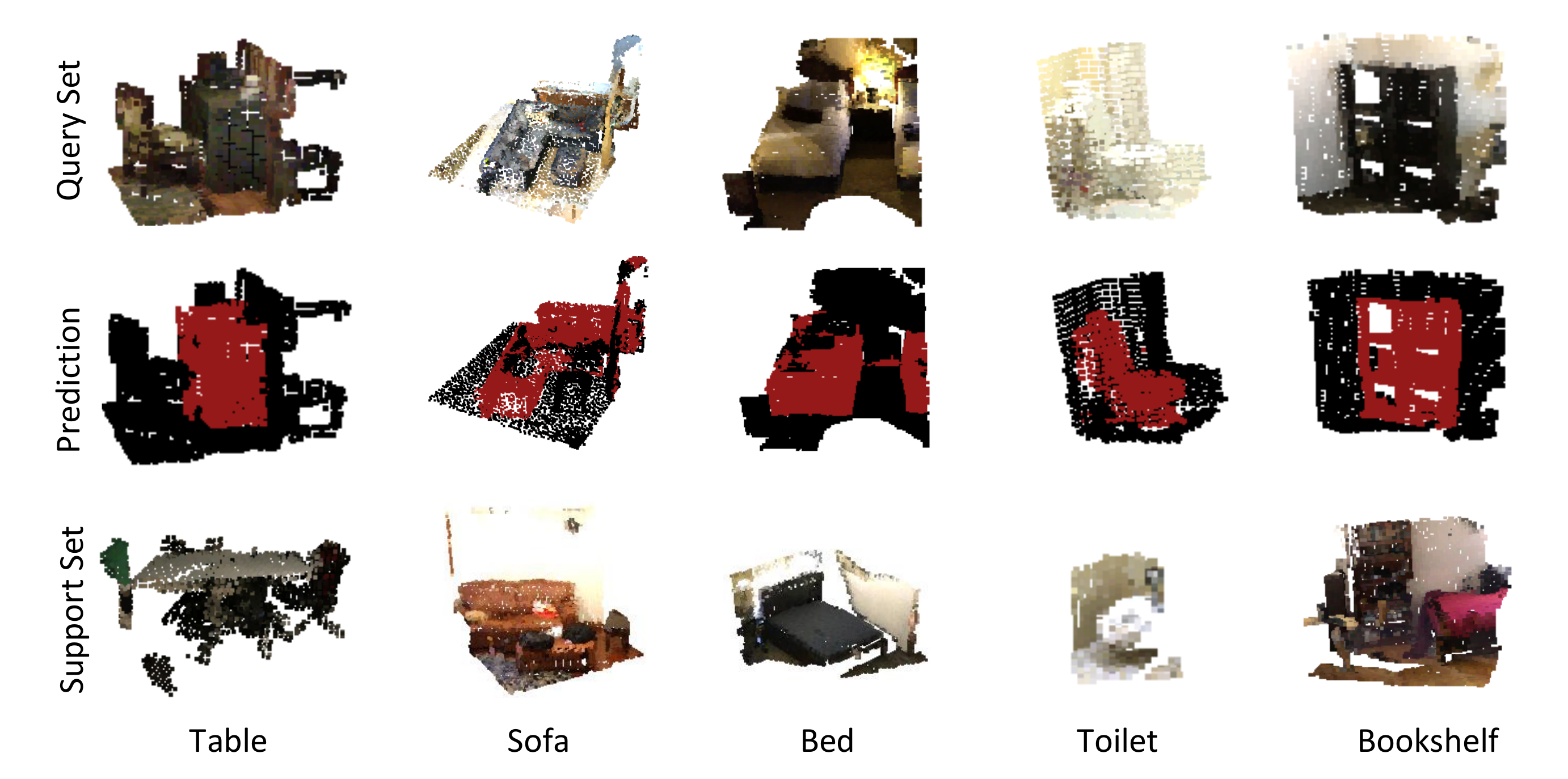}
  \caption{Visualization of the samples and results of one-shot segmentation on ScanNet-$6^{i}$, The proposed method is able to recognize the object from background with only a few samples.} .
  \label{fig:samples}
\end{figure}

\textbf{Dataset:} We create a new dataset, ScanNet-$6^{i}$, for the problem of $k$-shot point cloud Segmentation using point cloud data and annotations from ScanNet \citep{dai2017scannet}. As point clouds in ScanNet are large indoor scenes, the strong class imbalance exists, as shown in Table. \ref{fig:scannet_prop}. The imbalance makes the few-shot classes are easily ignored when evaluating in one-shot segmentation tasks, where all negative prediction gets high scores. We drop two categories (floor and wall), to maintain the rest 18 categories for the ScanNet-$6^{i}$. Besides, we crop the point clouds based on the instance annotations from ScanNet to balance the foreground and background points. Specifically, we crop a cube area around the geometry center for each instance; the cube size is twice bigger than the instance size. The samples of ScanNet-$6^{i}$ are shown in Fig.\ref{fig:samples}.

\begin{table*}[t]
\centering
\caption{Test classes for each fold and sample numbers for each class in ScanNet-${6^i}$. }
\resizebox{\textwidth}{!}{\begin{tabular}{|c|c|c|c|c|c|c|c|c|c|c|c|c|c|c|c|c|c|}
\hline
\multicolumn{6}{|c|}{$i = 0$} & \multicolumn{6}{|c|}{$i = 1$} & \multicolumn{6}{|c|}{$i = 2$} \\
\hline\hline
cabinet & bed & chair & sofa & table & door & window & bookshelf & picture & counter & desk & curtain & refrigerator & shower curtain & toilet & sink & bathtub & otherfurniture \\
\hline\
1799 & 388 & 5725 & 503 & 1621 & 2493 & 1210 & 377 & 883 & 268 & 678 & 359 & 243 & 144 & 259 & 488 & 144 & 2515 \\
\hline\end{tabular}}
\label{table:scannet_names}
\end{table*}

 The design and splits of categories for training and testing referenced the PASCAL-$5^i$ dataset \citep{shaban2017one}. We sample point cloud data with labels of six categories and consider them as the test set $Test^i$ = {6i, 6i+1, ..., 6i+5}, with i being the fold number and the rest data of 12 categories forming the training set $Train^i$. Test and training class names are shown in Table \ref{table:scannet_names}. The masks in $Train^i$ are modified so that any pixel with a category outside training categories is set as background. We follow a similar procedure to form the test set $Test^i$. We sample N=1000 examples and use them as the benchmark for evaluating the method described in the next section.

\textbf{Metric:} Given a set of predicted binary segmentation masks and the annotated mask for a semantic class $l$, we define the per-class Intersection over Union ($IoU_l$) as $\frac{\mathit{tp}_l}{\mathit{tp}_l + \mathit{fp}_l + \mathit{fn}_l}$.
Here, $\mathit{tp}_l$ is the number of true positives, $\mathit{fp}_l$ is the number of false positives and $\mathit{fn}_l$ is the number of false negatives over the set of masks. The meanIoU is its average over the set of classes.

In supervised segmentation experiments, we use the original ScanNet 3D Semantic label benchmark with the corresponding metrics to evaluate the proposed metric based supervised method for point cloud segmentation and compare our method with the state-of-art methods.

\begin{table}[t]
    \centering
    \caption{Comparison of Inner-Product, Convolution, and Multi-view Comparison (MVC) Component in the GAP experiment. Numbers in the brackets show the layer settings of the network, e.g., (64-16-1) means there is 3 layers, and the corresponding kernel numbers are 64, 16, and 1. The mIoU is gradually improved with the deeper MVC.}
    \resizebox{.53\textwidth}{!}{
    \begin{tabular}{l|c|r|r}
    \hline
    Methods& mIoU & MACS & Parameters \\
    \hline
        Inner Product & 33.8 & 1.28M & 0K \\
        Concat+Conv (64-16-1) & 38.1 & 175.10M &18K\\
        \hline
        MVC (1) & 34.0 & 1.30M & 16K\\
        MVC (8-1) & 35.4 & 10.44M &17K\\
        MVC (64-1) & 35.6 & 83.21M &25K\\
        MVC (64-16-1) & 38.9 & 93.12M &27K\\
    \hline
    \end{tabular}}
    \label{tab:classifier comp}
\end{table}

\begin{table}[t]
    \centering
    \caption{Comparison of Inner-Product Convolution and Multi-view Comparison (MVC) Component in the CP experiment. The MVC module has the advantages compared with the other popular similarity methods.}
    \resizebox{.27\textwidth}{!}{
    \begin{tabular}{l|c}
    \hline
    Methods& mIoU   \\
    \hline
        Inner-Product & 39.2 \\
        Concat+Conv & 39.7 \\
        MVC & 40.2 \\
    \hline
    \end{tabular}}
    \label{tab:gau comp}
\end{table}

\section{Baseline}

We set up two baseline networks for few-shot point cloud segmentation. One of the baselines uses global prototypes to compare between query and support set. The other baseline is to build a Compositional Prototype Network for better prototypes. We further enhance the baseline models with the proposed Multi-view Comparison Component.

The general framework consists of an encoder of Siamese Network, a relation module, and a classifier. We set up the baseline and proposed models with Submanifold Sparse Convolution \citep{3DSemanticSegmentationWithSubmanifoldSparseConvNet} for its high efficiency and low memory cost. The encoders are designed as Sparse Fully Convolutional Network(Sparse FCN), and classifiers are Sparse Residual layers. The main difference between our method and the baseline methods is the way to compare.

\begin{itemize}
\item \textbf{Global Averaged Prototype (GAP):} Relation Networks \citep{Sung_2018_CVPR} uses global average pooling to generate the global foreground prototype of support data. It then concatenates the prototype with the query features to obtain similarities by Convolution. 
\item \textbf{Compositional Prototypes (CP):} PGNet \citep{zhang2019pyramid} proposes Graph Attention Unit (GAU), which uses Graph Networks to compute attentive graph based on Inner-Product of each query point to obtain more accurate prototype for each query point. In the experiment, We reference the Graph Attention Unit (GAU) of PGNet to build our relation module and use a convolutional model to calculate similarities in our baselines.

\end{itemize}

\begin{table*}[t]
\centering
\caption{Mean IoU results  on all folds of ScanNet-$6^i$ for the $1$-shot and $5$-shot tasks respectively. GAP denoted the Global Averaged Prototype Network, CP is Compositional Prototype Network, and CP with MVC is the improved Compositional Prototype Network based on Multi-view Comparison Component (MVC).}
\resizebox{.65\textwidth}{!}{\begin{tabular}{c|cccc||c}\\
{\bf Methods ($1$-shot)} & ScanNet-$6^0$ & ScanNet-$6^1$ & ScanNet-$6^2$ & Mean\\
\hline\hline
GAP & $41.6$ & $30.1$ & $37.4$ & $36.4$ \\
CP & $41.9$ & $29.0$ & $36.6$ & $35.8$ \\
CP with MVC & $\mathbf{43.4}$ & $\mathbf{31.7}$ & $\mathbf{38.1}$ & $\mathbf{37.7}$ \\
\end{tabular}}
\centering
\resizebox{.65\textwidth}{!}{\begin{tabular}{c|cccc||c}\\
{\bf Methods ($5$-shot)} & 
ScanNet-$6^0$ & ScanNet-$6^1$ & ScanNet-$6^2$ & Mean\\
\hline\hline
GAP & $43.9$ & $35.4$ & $43.1$ & $40.8$\\
CP & $44.3$ & $35.3$ & $41.2$ & $40.3$\\
CP with MVC & $\mathbf{45.1}$ & $\mathbf{35.8}$ & $\mathbf{44.5}$ & $\mathbf{41.8}$\\
\end{tabular}}
\label{table:full_results}
\end{table*}

\section{Experiments}

\subsection{Ablation Study}

In this section, We set up experiments to study the proposed Multi-view Comparison (MVC) Component of different settings, and combine them with the baseline methods. The experiments are based on the split of ScanNet-$6^0$.

We evaluate the similarity model of MVC, Convolution, and Inner-Product in the global averaged prototype (GAP) experiment. When the numbers of layers and kernels are set to 1, the MVC is similar to Inner-Product. The only layer helps MVC adapts its kernel to query points. We gradually add views and layers in MVC, and The results are shown in Table. \ref{tab:classifier comp} The Multiply Accumulates (MACs) are measured on 10000 features of points. MVC has advantages than other metrics. MVC also has fewer MACs than the Convolution method, where input features are concatenated before Convolution in the first layer. The multi-view Kernel generation in MVC also involves extra computation, but the increase is negligible.

In the compositional prototypes (CP) experiment, We adopt the Graph Attention Unit (GAU) to improve the prototypes. Convolution is used to calculate the relation between compositional prototypes and query features for all models. We promote the Inner-Product in original GAU with Multi-view Comparison Component and Convolution method. The layer settings of Relation Network and MVC are set to 3 layers with corresponding kernel numbers of 64, 16, 1. The results in Table. \ref{tab:gau comp} also shows the advantages of MVC.

\subsection{One-Shot Segmentation Experiments on ScanNet-$6^i$}

In this section, we conduct $k$-shot experiments to evaluate the performance of our method and the baseline methods. The 1-shot and 5-shot results are shown in Table. \ref{table:full_results}. On both $1$-shot and $5$-shot tasks, the improved CP methods with MVC has the best results. The original CP Network does not show the advantages compared with the Global Averaged Prototype (GAP) method. Compared with the 1-shot task, the 5-shot task provides more support data, which helps the models to compute more general prototypes, and all models obtain better results.

\subsection{Supervised Segmentation Experiments Based on Metric Learning}

\begin{table}[t]
    \centering
    \caption{3D Semantic Label Benchmark on ScanNet  \citep{dai2017scannet}. The proposed method improved the performance on the few-shot classes and the overall results.}
    \resizebox{.5\textwidth}{!}{
    \begin{tabular}{r|c|c|c|c}
    \hline
            Method  & Few & Medium & Many & Overall \\
    \hline\hline
        MVPNet  \citep{jaritz2019multi} & $55.0$ & $64.1$ & $86.5$ & $64.1$ \\
        \hline
        PointConv~ \citep{wu2019pointconv} & $62.3$ & $66.8$ & $88.3$ & $66.6$ \\
        KP-FCNN~ \citep{thomas2019kpconv} & $63.8$ & $67.8$ & $87.7$ & $68.4$ \\
    \hline
        SparseConvNet~ \citep{3DSemanticSegmentationWithSubmanifoldSparseConvNet} & $67.8$ & $\mathbf{72.1}$ & $\mathbf{91.0}$ & $72.5$ \\
        Ours  & $\mathbf{69.7}$ & $71.3$ & $90.5$ & $\mathbf{72.7}$ \\
    \hline
    \end{tabular}}
    \label{tab:scannet}
\end{table}

We design a metric-based method in supervised point cloud segmentation, as mentioned in section \ref{sup}, to help the model to learn balanced representations on the class-imbalanced dataset. We initial prototypes for each class and modify the classifier of the Submanifold Sparse Convolutional Network (SparseConvNet) \citep{3DSemanticSegmentationWithSubmanifoldSparseConvNet} to compare the embeddings with the prototypes for classification based on the Multi-view Comparison Component. 

We evaluate the performance on few-shot, medium-shot, and many-shot classes defined in Table. \ref{fig:scannet_prop}. We do not perform oversampling or re-weighting of the rare class examples during training. From the results in Table. \ref{tab:scannet}, the proposed method enhances the performance of few-shot classes and has advantages on overall performance.

\section{Conclusion}

In this paper, we have presented a novel Multi-view Comparison Component, which has advantages on efficiency and performance. We have designed the Compositional Prototype Network for few-shot point cloud segmentation. In the supervised task, our metric-based method also alleviates the class-imbalance effect. Experiments on ScanNet and ScanNet-$6^{i}$ show the clear superiority of our methods. 

\section*{Broader Impact}

There are no ethical and future societal consequences of our work.

\medskip

\small
\bibliographystyle{unsrt}
\bibliography{egbib}

\begin{thebibliography}{10}

\bibitem{dai2017scannet}
Angela Dai, Angel~X. Chang, Manolis Savva, Maciej Halber, Thomas Funkhouser,
  and Matthias Nie{\ss}ner.
\newblock Scannet: Richly-annotated 3d reconstructions of indoor scenes.
\newblock In {\em Proc. Computer Vision and Pattern Recognition (CVPR), IEEE},
  2017.

\bibitem{Sung_2018_CVPR}
Flood Sung, Yongxin Yang, Li~Zhang, Tao Xiang, Philip~H.S. Torr, and Timothy~M.
  Hospedales.
\newblock Learning to compare: Relation network for few-shot learning.
\newblock In {\em The IEEE Conference on Computer Vision and Pattern
  Recognition (CVPR)}, June 2018.

\bibitem{wang2019panet}
Kaixin Wang, Jun~Hao Liew, Yingtian Zou, Daquan Zhou, and Jiashi Feng.
\newblock Panet: Few-shot image semantic segmentation with prototype alignment.
\newblock In {\em Proceedings of the IEEE International Conference on Computer
  Vision}, pages 9197--9206, 2019.

\bibitem{liu2020crnet}
Weide Liu, Chi Zhang, Guosheng Lin, and Fayao Liu.
\newblock Crnet: Cross-reference networks for few-shot segmentation.
\newblock In {\em Proceedings of the IEEE/CVF Conference on Computer Vision and
  Pattern Recognition}, pages 4165--4173, 2020.

\bibitem{zhang2019canet}
Chi Zhang, Guosheng Lin, Fayao Liu, Rui Yao, and Chunhua Shen.
\newblock Canet: Class-agnostic segmentation networks with iterative refinement
  and attentive few-shot learning.
\newblock In {\em Proceedings of the IEEE Conference on Computer Vision and
  Pattern Recognition}, pages 5217--5226, 2019.

\bibitem{zhang2019pyramid}
Chi Zhang, Guosheng Lin, Fayao Liu, Jiushuang Guo, Qingyao Wu, and Rui Yao.
\newblock Pyramid graph networks with connection attentions for region-based
  one-shot semantic segmentation.
\newblock In {\em Proceedings of the IEEE International Conference on Computer
  Vision}, pages 9587--9595, 2019.

\bibitem{shaban2017one}
Amirreza Shaban, Shray Bansal, Zhen Liu, Irfan Essa, and Byron Boots.
\newblock One-shot learning for semantic segmentation.
\newblock {\em arXiv preprint arXiv:1709.03410}, 2017.

\bibitem{Zhang_2019_ICCV}
Chi Zhang, Guosheng Lin, Fayao Liu, Jiushuang Guo, Qingyao Wu, and Rui Yao.
\newblock Pyramid graph networks with connection attentions for region-based
  one-shot semantic segmentation.
\newblock In {\em The IEEE International Conference on Computer Vision (ICCV)},
  October 2019.

\bibitem{guerriero2018deep}
Samantha Guerriero, Barbara Caputo, and Thomas Mensink.
\newblock Deep nearest class mean classifiers.
\newblock In {\em International Conference on Learning Representations,
  Worskhop Track}, 2018.

\bibitem{kang2019decoupling}
Bingyi Kang, Saining Xie, Marcus Rohrbach, Zhicheng Yan, Albert Gordo, Jiashi
  Feng, and Yannis Kalantidis.
\newblock Decoupling representation and classifier for long-tailed recognition.
\newblock {\em arXiv preprint arXiv:1910.09217}, 2019.

\bibitem{snell2017prototypical}
Jake Snell, Kevin Swersky, and Richard Zemel.
\newblock Prototypical networks for few-shot learning.
\newblock In {\em Advances in neural information processing systems}, pages
  4077--4087, 2017.

\bibitem{vinyals2016matching}
Oriol Vinyals, Charles Blundell, Timothy Lillicrap, Daan Wierstra, et~al.
\newblock Matching networks for one shot learning.
\newblock In {\em Advances in neural information processing systems}, pages
  3630--3638, 2016.

\bibitem{zhang2020deepemd}
Chi Zhang, Yujun Cai, Guosheng Lin, and Chunhua Shen.
\newblock Deepemd: Few-shot image classification with differentiable earth
  mover's distance and structured classifiers.
\newblock In {\em Proceedings of the IEEE/CVF Conference on Computer Vision and
  Pattern Recognition}, pages 12203--12213, 2020.

\bibitem{bertinetto2016fully}
Luca Bertinetto, Jack Valmadre, Jo{\~a}o~F Henriques, Andrea Vedaldi, and
  Philip H~S Torr.
\newblock Fully-convolutional siamese networks for object tracking.
\newblock In {\em ECCV 2016 Workshops}, pages 850--865, 2016.

\bibitem{wang2019fast}
Qiang Wang, Li~Zhang, Luca Bertinetto, Weiming Hu, and Philip~HS Torr.
\newblock Fast online object tracking and segmentation: A unifying approach.
\newblock In {\em Proceedings of the IEEE conference on computer vision and
  pattern recognition}, pages 1328--1338, 2019.

\bibitem{qi2017pointnet}
Charles~R Qi, Hao Su, Kaichun Mo, and Leonidas~J Guibas.
\newblock Pointnet: Deep learning on point sets for 3d classification and
  segmentation.
\newblock In {\em Proceedings of the IEEE conference on computer vision and
  pattern recognition}, pages 652--660, 2017.

\bibitem{qi2017pointnet++}
Charles~Ruizhongtai Qi, Li~Yi, Hao Su, and Leonidas~J Guibas.
\newblock Pointnet++: Deep hierarchical feature learning on point sets in a
  metric space.
\newblock In {\em Advances in neural information processing systems}, pages
  5099--5108, 2017.

\bibitem{thomas2019kpconv}
Hugues Thomas, Charles~R Qi, Jean-Emmanuel Deschaud, Beatriz Marcotegui,
  Fran{\c{c}}ois Goulette, and Leonidas~J Guibas.
\newblock Kpconv: Flexible and deformable convolution for point clouds.
\newblock In {\em Proceedings of the IEEE International Conference on Computer
  Vision}, pages 6411--6420, 2019.

\bibitem{jaritz2019multi}
Maximilian Jaritz, Jiayuan Gu, and Hao Su.
\newblock Multi-view pointnet for 3d scene understanding.
\newblock In {\em Proceedings of the IEEE International Conference on Computer
  Vision Workshops}, pages 0--0, 2019.

\bibitem{3DSemanticSegmentationWithSubmanifoldSparseConvNet}
Benjamin Graham, Martin Engelcke, and Laurens van~der Maaten.
\newblock 3d semantic segmentation with submanifold sparse convolutional
  networks.
\newblock {\em CVPR}, 2018.

\bibitem{wu2019pointconv}
Wenxuan Wu, Zhongang Qi, and Li~Fuxin.
\newblock Pointconv: Deep convolutional networks on 3d point clouds.
\newblock In {\em Proceedings of the IEEE Conference on Computer Vision and
  Pattern Recognition}, pages 9621--9630, 2019.

\end{thebibliography}

\end{document}